\pdfoutput=1

\documentclass[11pt]{article}

\usepackage[preprint]{acl}

\usepackage{times}
\usepackage{latexsym}

\usepackage[T1]{fontenc}

\usepackage[utf8]{inputenc}

\usepackage{microtype}

\usepackage{inconsolata}
\usepackage{graphicx}
\usepackage{multirow}
\usepackage{ltablex}

\title{FoundaBench: Evaluating Chinese Fundamental Knowledge Capabilities of Large Language Models}

\author{Wei Li, Ren Ma, Jiang Wu, Chenya Gu, Jiahui Peng, Jinyang Len \\
    \textbf{Songyang Zhang}, \textbf{Hang Yan},  \textbf{Dahua Lin}, \textbf{Conghui He}\thanks{Corresponding author}  \\
Shanghai Artificial Intelligence Laboratory\\
Shanghai, 200232, China\\
    \texttt{heconghui@pjlab.org.cn} \\}
    
\begin{document}
\maketitle
\begin{abstract}

In the burgeoning field of large language models (LLMs), the assessment of fundamental knowledge remains a critical challenge, particularly for models tailored to Chinese language and culture. This paper introduces FoundaBench, a pioneering benchmark designed to rigorously evaluate the fundamental knowledge capabilities of Chinese LLMs. FoundaBench encompasses a diverse array of 3354 multiple-choice questions across common sense and K-12 educational subjects, meticulously curated to reflect the breadth and depth of everyday and academic knowledge. We present an extensive evaluation of 12 state-of-the-art LLMs using FoundaBench, employing both traditional assessment methods and our CircularEval protocol to mitigate potential biases in model responses. Our results highlight the superior performance of models pre-trained on Chinese corpora, and reveal a significant disparity between models' reasoning and memory recall capabilities. The insights gleaned from FoundaBench evaluations set a new standard for understanding the fundamental knowledge of LLMs, providing a robust framework for future advancements in the field.

\end{abstract}

\section{Introduction}
As the field of large language models (LLMs) progresses, there has been a noticeable escalation in their quantity, parameter magnitude, and overall capabilities. While certain large language models possess the capacity to store and recall extensive world knowledge, they still exhibit a discernible gap when compared to the basic knowledge capabilities of ordinary individuals. The knowledge requisites of a large language model can be divided into three categories: fundamental knowledge, specialized knowledge, and time-relevant knowledge. 

Specialized knowledge refers to specific industries or domains and is in-depth and often confined to its field of application. A general model need not master all such domains, since this, along with time-relevant knowledge, can be sourced from knowledge bases. In contrast, fundamental knowledge, which includes everyday life experiences and basic theoretical knowledge, is universal and will not change over time. 

Parallel to the rapid advancement of Chinese LLMs, there has been a significant increase in the emergence of Chinese benchmarks. These can be principally divided into two primary categories. The first category focuses on common sense or world knowledge, many translations of English benchmarks, such as XCOPA \cite{ponti-etal-2020-xcopa} and XStoryCloze \cite{lin2022fewshot}. However, these often lack comprehensive applicability to Chinese culture. The second category includes subject-oriented benchmarks such as GAOKAO-Bench \cite{zhang2024evaluating}, AGIEval \cite{zhong2023agieval} and Xiezhi \cite{gu2023xiezhi}. These benchmarks may lack a distinction between the difficulty of the evaluation and may not adequately correspond to the level of fundamental knowledge of ordinary people.

\begin{figure*}
    \centering
    \includegraphics[width=1.0\textwidth]{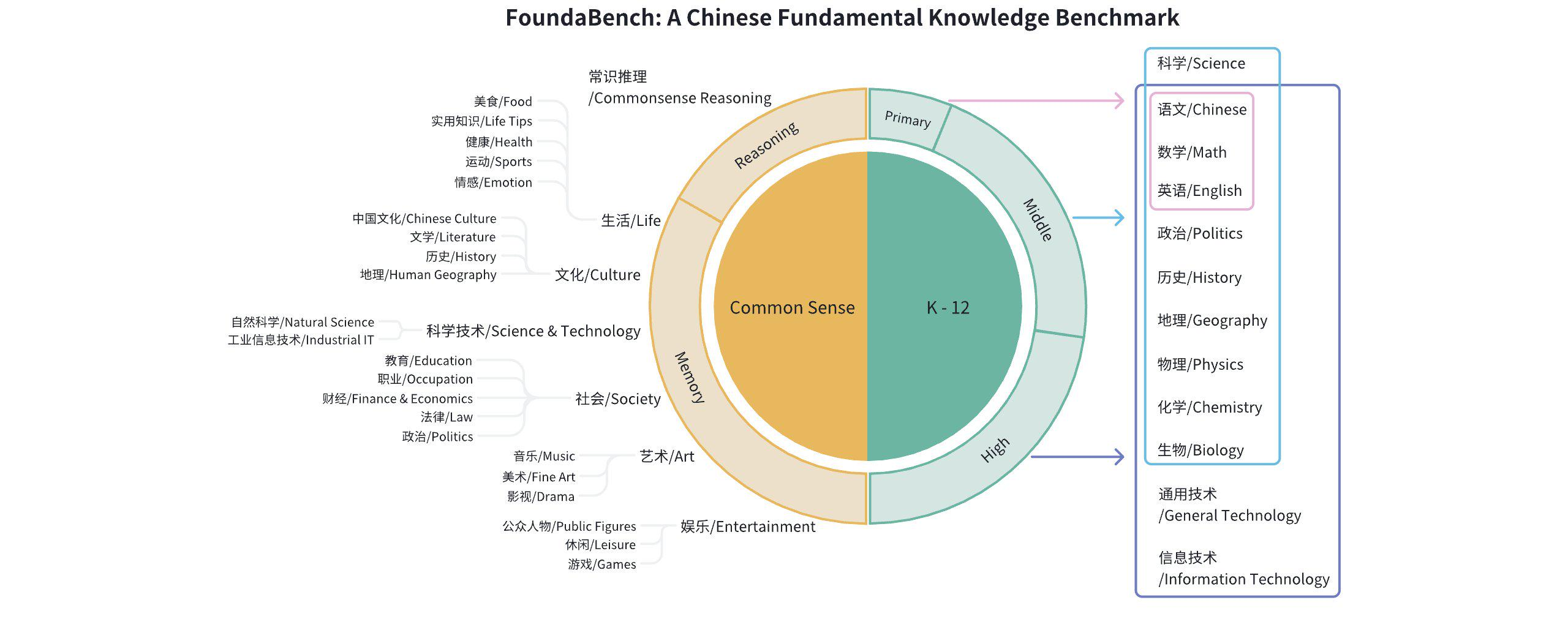}
    \caption{Overview diagram of the FoundaBench}
    \label{fig:figure1}
\end{figure*}

In this paper, we present FoundaBench, the first comprehensive benchmark specifically designed to evaluate the fundamental knowledge capabilities of Chinese LLMs. This benchmark is constructed based on a taxonomy tailored for fundamental knowledge, divided into two sections: common sense and K-12 subjects. The questions are derived from a wide range of sources, including Wikipedia, various knowledge platforms, and Chinese educational platforms available on the Internet. FoundaBench includes 3354 multiple choice questions that cover everyday common sense reasoning and knowledge, including Chinese culture, society, art, entertainment, and also extend to the K-12 educational stages, including primary, middle, and high school. The taxonomy is depicted in Figure~\ref{fig:figure1}.

We collected original evaluation questions and answers from various sources and performed a thorough cleaning and processing. We also collect textbooks, documents that include knowledge points, and general knowledge to design an optimized prompting process, thus generating and improving evaluation questions to ensure their difficulty level. Furthermore, we incorporate psychostatistical methods to assess and improve the reliability and validity of the entire evaluation set, ensuring its effectiveness. We also balanced the number of questions across each domain and category to align with the fundamental knowledge capabilities of an ordinary person.

Furthermore, we evaluated 12 models of varying sizes and language orientations in a normal evaluation method and 7 of them in CircularEval\cite{liu2023mmbench} to reveal the exact results of the evaluations. This evaluation aimed to experiment with current top-performing models to evaluate the level of fundamental knowledge capabilities and identify the influencing factors of them.

In summary, the contributions of our work are as follows:
\begin{itemize}
\item We construct FoundaBench and propose a taxonomy for evaluating the fundamental knowledge capabilities of large language models, covering common sense and K-12 educational subjects. The benchmark is comprehensive and covers five main categories, 19 subjects, and 35 subcategories, including 1529 knowledge points.

\item In terms of quality assessment of the benchmark questions, we have defined detailed multidimensional quality standards for the content. This, in combination with manual quality control, ensures consistency with the level of fundamental human knowledge capabilities by introducing the reliability and validity concepts of psychostatistical methods.

\item We employ evaluation methods that include zero-shot and circular comparisons. We perform analysis on 12 different open-source and closed-source models with varying parameter scales, thus defining a baseline for the fundamental knowledge capabilities of LLMs.
\end{itemize}

\section{Related work}

\subsection{Common sense benchmarks}
Regarding the Chinese common sense benchmarks, there are two categories. The first category is the benchmarks that are originally English and translated to Chinese, such as XCOPA \cite{ponti-etal-2020-xcopa} and XStoryCloze \cite{lin2022fewshot}. However, these translated benchmarks lack Chinese factors, such as culture, region, history, and lifestyle habits, and therefore cannot fully and comprehensively assess the performance of LLMs in the Chinese scenario. 

The second category is the native Chinese benchmarks. The elementary common sense task in CMMLU mainly focuses on topics such as heatstroke, fire, diet, first aid, etc. for elementary students. CORECODE \cite{shi2023corecode} is the dataset aimed at evaluating common sense conflict detection in Chinese LLMs. Some benchmarks are focused on one topic, such as RoleEval-Chinese \cite{shen2023roleeval}, which is used to evaluate the capabilities related to famous Chinese characters. 

However, all the above benchmarks are not comprehensive and complete enough for the Chinese scenario, and the data collection and curation process is not systematic and lacks a reasonable classification system.

\subsection{K-12 subjects benchmarks}
K-12 education is an important educational phase in China that refers to the sum of primary and secondary education. Almost all Chinese people had K-12 education, so the subjective knowledge they learned in K-12 education should be the fundamental and general knowledge of Chinese LLMs.

Previous Chinese benchmarks, including CEval \cite{huang2023ceval}, CMMLU \cite{li2024cmmlu}, GAOKAO-Bench \cite{zhang2024evaluating}, AGIEval \cite{zhong2023agieval}, and Xiezhi \cite{gu2023xiezhi}, have also included subjects at the K-12 level. Among these, only Xiezhi includes subjects at the elementary school level. Unfortunately, we cannot isolate questions specifically relevant to the elementary school level due to the lack of grade information provision by Xiezhi. This presents a challenge in curating content that is most appropriate for students at this particular academic stage. Furthermore, the lack of information about key knowledge points in these benchmarks undermines the granularity of insights that can be derived from benchmark scores.

\section{Chinese Fundamental Knowledge Benchmark}
\subsection{Design principle}
\textbf{Overview. }
Before starting to describe the design of a taxonomy, it is crucial to establish a clear definition of 'fundamental knowledge'.  We define fundamental knowledge as the knowledge that a person of normal intelligence in society should or would have. This encompasses both common sense, often referred to everyday life experience, and subject-specific knowledge typically acquired through education.

\begin{table}[ht]
\centering
\begin{tabular}{llc}
\hline
\textbf{Category} & \textbf{Subjects} & \textbf{Number of}\\
 &   & \textbf{ Questions}\\
\hline
Common & Reasoning & 289 \\
Sense & Life & 111 \\
& Culture & 264 \\
& Society & 128 \\
& Sci. \& Tech. & 96 \\
& Art & 54 \\
& Entertainment & 43 \\
K-12 & Chinese & 291 \\
& Math & 295 \\
& English & 300 \\
& Politics & 189 \\
& History & 200 \\
& Geography & 200 \\
& Physics & 200 \\
& Chemistry & 199 \\
& Biology & 200 \\
& Science & 96 \\
& General Tech. & 100 \\
& Info. Tech. & 99 \\
Total & & 3354 \\
\hline
\end{tabular}
\caption{Statistics of FoundaBench}
\label{tab:statistic1}
\end{table}

The taxonomy for foundational knowledge in LLMs aligns with human learning processes, drawing from everyday experiences and academic knowledge from compulsory education. Therefore, the dataset is divided into a common sense subset and a subject-specific subset for grades K-12. However, the data handling methods differ from source to source. A comprehensive breakdown of subjects and their categories is provided in Table~\ref{tab:statistic1}.

\textbf{Taxonomy design. }
In accordance with our design principles, the taxonomy can be divided into two sections: common sense and K-12 subjects. The common sense taxonomy is derived from a comprehensive analysis of various sources, including the Chinese version of Wikipedia, leading Chinese knowledge platforms such as Zhihu and Baidu Tieba, and popular topics from online LLM chat platforms. This ensures coverage of the categories of information that are most interesting to people. Referring to the above categories, we combined similar categories and the taxonomy designed as in Figure~\ref{fig:figure1}.

\subsection{Data construction}
The dataset curation and dataset construction pipeline is introduced as follows. 

\textbf{Data collection. }
Drawing on the common sense section of our taxonomy, we collected questions from the Internet, including those from Chinese civil service exams. These exams test candidates' common and basic knowledge and their analytical abilities, mirroring our objective of evaluating LLMs. Additionally, we incorporate web-curated questions from online users, focusing on traditional Chinese culture and life experiences.

For the K-12 section, we gather questions from Chinese academic examinations, targeting specific disciplines such as mathematics, science, and history. To standardize future evaluations, we prioritized multiple-choice questions with a single correct answer, referred to as closed questions.

Next, we categorized all the questions according to our taxonomy, supplementing uncovered areas like entertainment, life, and art knowledge. We aimed to find some documents and textbooks summarizing general knowledge, laying the groundwork for future semiautomatic question generation.

\textbf{Data cleaning and processing. }
The original questions, arriving in various formats, were converted into a structured layout, with those offering non-four-option answers or multiple correct answers removed. Knowledge summary documents were refined by excluding lengthy explanations and keeping simplified concept or noun descriptions. Content with errors or less common concepts was manually removed.

To adhere to our text-only benchmark, questions containing images were excluded. Each question was structured to include a unique ID, description, four choices, a standard answer, and classification.

We then de-duplicated the multiple-choice questions to avoid repetition. These questions were manually labeled and filtered, excluding those with problematic phrasing, incorrect answers, or non-commonsense judgments. The remaining questions were classified according to our taxonomy.

\textbf{Automatic generation. }
While GPT is commonly used for question testing, we have ventured into its application to generate evaluation questions. Our initial step involved extracting key content from knowledge documents and formulating a prompt using collected course outlines, test points, and the corresponding examples, and then using GPT4 to generate sample questions.

\begin{figure*}
    \centering
    \includegraphics[width=0.9\linewidth]{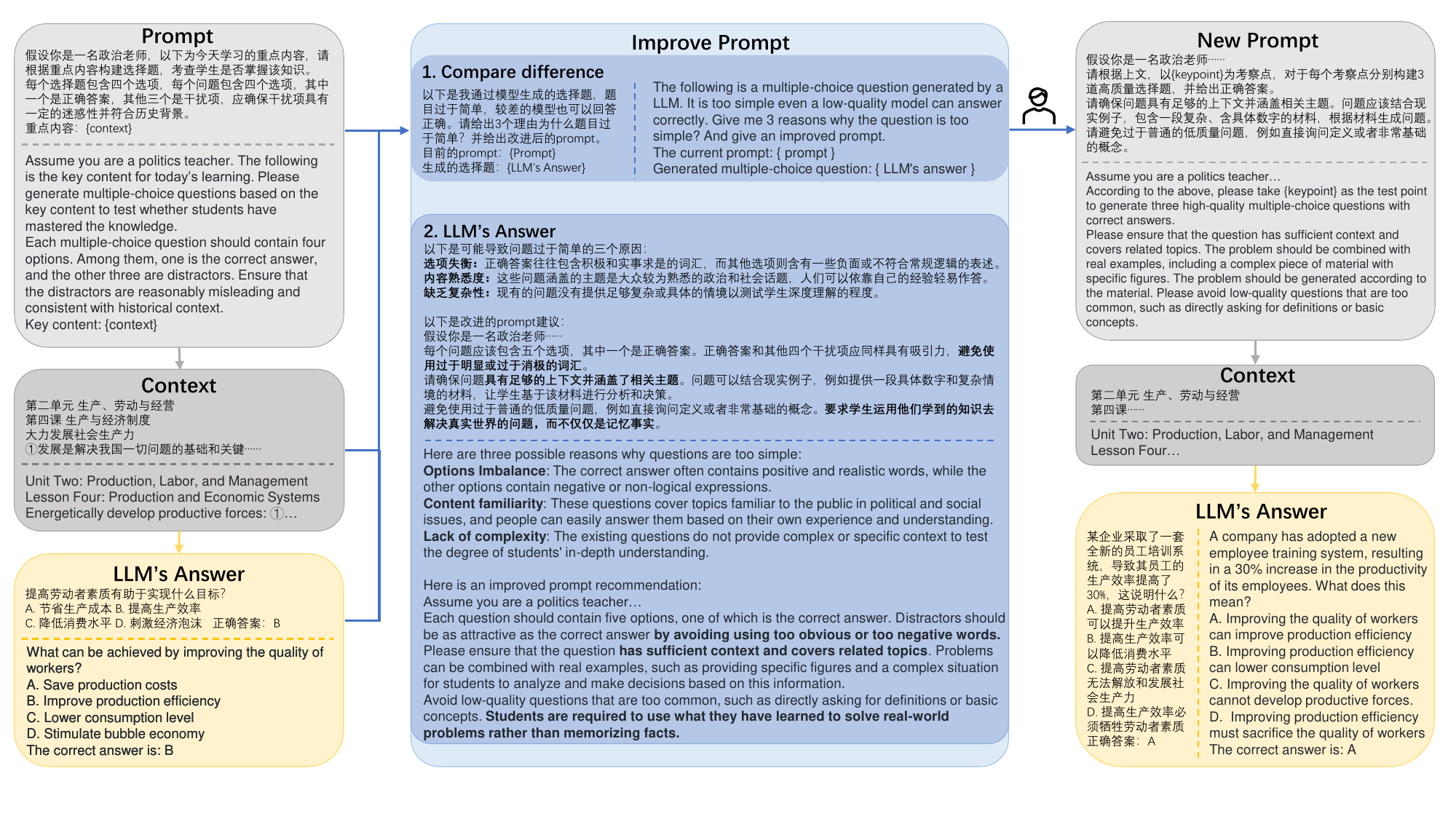}
    \caption{Improved process for question generation}
    \label{fig:figure2}
\end{figure*}

However, the initial output revealed issues such as inadequate complexity, lack of specific corpus integration, and unbalanced options. To rectify these, we sent the initial questions back to GPT4 for further questioning, leading to improved prompts. These prompts were manually verified, and unsuitable ones were discarded, resulting in a refined new prompt. For instance, we introduced a requirement for five-option questions to enhance complexity. Using this refined prompt, we generated more context-specific questions. This optimization process is depicted in Figure~\ref{fig:figure2}.

The provision of content in batches facilitated the generation of hundreds of questions, encompassing a wide range of categories such as traditional Chinese culture, art and entertainment, life and health, and the majority of K-12 subjects. 

\subsection{Quality control}
In the quality control process, we innovatively incorporate psychostatistical methods to ensure the quality of the benchmark dataset. This creative approach represents a departure from methods traditionally employed in previous benchmark curation.

\textbf{Align with human level. }
Given that our benchmark dataset is designed to evaluate LLMs' comprehension and application of fundamental knowledge derived from the human world, human judgment shall be involved to determine whether these questions indeed represent fundamental knowledge.

According to the definition of common sense, a question is not considered common sense if most people cannot answer it correctly based on their intuition and the existing long-term memory. Therefore, if individuals can answer common sense questions in our benchmark dataset with high accuracy, we regard the dataset as high quality.

In the K-12 section, we consider the quality of the questions to be high when they meet the standards of Chinese primary and secondary school exams, and the evaluation set covers most of the knowledge points. The statistics of the key points are provided in Table~\ref{tab:statistic2}.

\begin{table}[ht]
\centering
\begin{tabular}{lc}
\hline
\textbf{Subjects} & \textbf{Number of }\\
 & \textbf{Key points}\\
\hline
Chinese & 153 \\
Math & 238 \\
English & 166 \\
Politics & 122 \\
History & 133 \\
Geography & 121 \\
Physics & 154 \\
Chemistry & 140 \\
Biology & 135 \\
Science & 70 \\
General Technology & 51 \\
Information Technology & 51 \\
Total & 1529\\
\hline
\end{tabular}
\caption{Statistics of key points}
\label{tab:statistic2}
\end{table}

\textbf{Model-based quality assurance. }
Before considering whether a question qualifies as common sense or meets the requirements of a standard test, the question itself must be correct. This implies that the phrasing of the question is accurate and unambiguous, there are no erroneous characters, there is a strong correlation among the choices provided within a question, and the answers are correct.

We initially assessed the quality of the questions and their answers, revising or discarding those of low quality. This process used both automated corrections by InternLM developed by Shanghai AI Lab\cite{2023internlm} and few manual adjustments. Automated corrections is realized by constinuously optimizing the prompt and have the model modify the questions and answers in the dataset in batches. Questions with numerous errors in their statements or answers were eliminated.

\textbf{Psychostatistical methods. }
Then we employed psychostatistical methods to improve the quality of the entire dataset. Psychological scales, frequently used in psychological research to assess individual abilities, rely on two important indices for quality evaluation: reliability and validity. We treated our benchmark dataset as such a scale, designed to gauge the capability of the model, thereby necessitating the enhancement of its reliability and validity.

\textbf{Reliability assurance. }
Reliability refers to the stability and consistency of the measurement results of a scale, and we tested it by evaluating multiple models twice on the same dataset. Variance in responses could be due to ambiguous questions, non-unique answers, or model instability. Despite these factors, the high correlation between two sets of results represented a reliable, high-quality evaluation dataset.

\textbf{Validity assurance. }
Validity, refers to accuracy and relevance of a scale, was ensured by the evaluation of experts and the accuracy calculations of the annotators using our dataset. Given the varying difficulty levels of the questions, we expected the accuracy to follow a normal distribution. Questions with annotator accuracy less than two standard deviations were removed as non-commonsense questions, enhancing the dataset's validity and its effectiveness in evaluating the fundamental knowledge capabilities of LLMs.

\begin{figure*}
    \centering
    \includegraphics[width=0.9\linewidth]{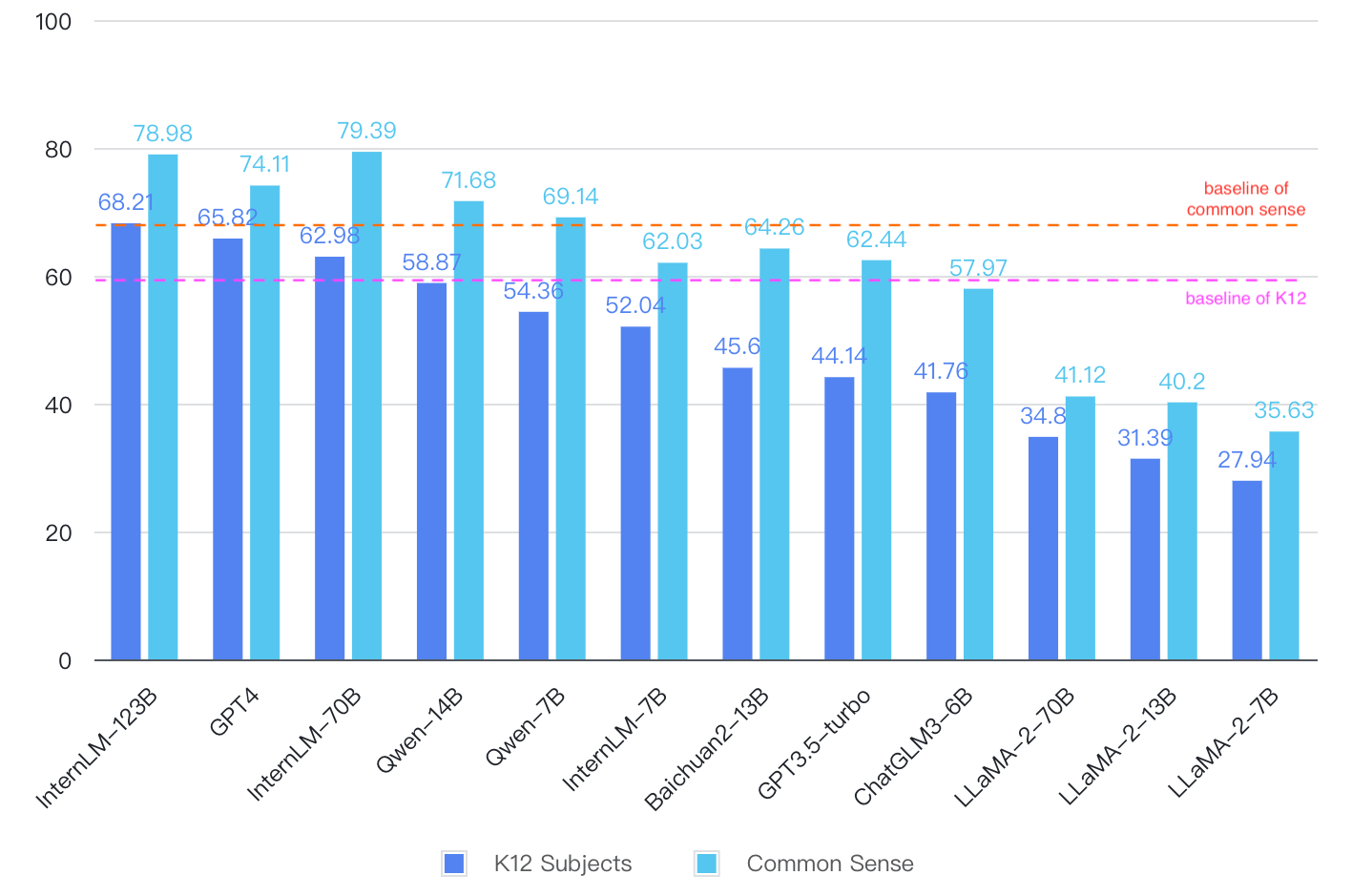}
    \caption{Evaluation of all models}
    \label{fig:evaluation-all}
\end{figure*}

\section{Experiments}
In this section, we evaluate the performance of various state-of-the-art language models on our benchmark dataset. We selected models that encompass various sizes and language orientations, but all in fine-tuned stages. We scored 12 models with the normal evaluation method and 7 of them with the CircularEval method. We compare their answers against the standard answers across all examples and tasks. Detailed descriptions of the evaluation results are listed in Appendix~\ref{sec:appendixA}.

\subsection{Evaluation}
\textbf{Assessment. }
First, we evaluate the models' answers with the original choice sequence, in the same way as a human performing exams. While in the context of multiple choice questions, a correct response from a language model does not necessarily denote a genuine comprehension and reasoning of the question; it could be a guess. As indicated by OpenCompass\cite{2023opencompass}, the CircularEval methodology can be used to distinguish these situations and mitigate the bias of language models towards certain options. This approach involves enhancing the questions by shifting its options. If the LLM correctly answers all variations of the augmented question, it is considered correct under the CircularEval framework. In our research, we implemented this methodology in a zero-shot setting and the evaluation results using CircularEval are listed in the Appendix~\ref{sec:appendixA}.

\textbf{Prompt. }
We employ a prompt, serving as the template for the test, detailed as: 
\begin{verbatim}
Question: {question}
A. {textA}
B. {textB}
C. {textC}
D. {textD}
Answer: {answer}
\end{verbatim} 
In the case of zero-shot evaluation, We present the questions directly in the prompt. 

\subsection{Models}
We access 12 top-performing LLMs in different sizes from 6 model families. As listed in Figure~\ref{fig:evaluation-all}, to provide an overview of the existing LLMs in the dataset, we selected both English-oriented and Chinese-oriented models.

GPT-4 and GPT-3.5 are the strongest generation models from OpenAI, delivering powerful performance across a range of languages. We employed versions "GPT-4-1106-preview" and "GPT-3.5-turbo-0613" in our research. LLaMA2, another well-regarded model, delivers strong results with limited data, primarily trained on English corpora from diverse sources. 

Additionally, Chinese-oriented models including InternLM, Qwen, Baichuan, and ChatGLM are also selected to evaluate, which are further adapted on conversational data. InternLM \cite{2023internlm} is developed by Shanghai AI Lab, which is an open-source lightweight training model that pre-trains on data with over a trillion tokens. Qwen \cite{bai2023qwen}, open-sourced by Alibaba, pretrained on 22 trillion tokens, supporting English and Chinese. Baichuan2 \cite{baichuan2023baichuan2} is the new generation of large language model launched by Baichuan Intelligent Technology, which is pretrained on high-quality corpora with 2.6 trillion tokens. ChatGLM \cite{zeng2022glm} is jointly released by Zhipu AI and Tsinghua KEG, supporting code execution, tool invocation and agent tasks in complex scenarios.

\subsection{Results and analysis}

After verifying the validity of the scoring using FoundaBench, we analyzed the level of fundamental knowledge of models, the effect of the evaluation methods, and hard examples and obtained the following conclusions.

\textbf{Level of fundamental knowledge. }
Among the comprehensive scores of all models, InternLM-123B and GPT4 have the highest scores, indicating that they have the best effect on mastering the basic knowledge of Chinese. Besides the effect of parameters' size, InternLM's series is even better than the GPT series. This is due to the pre-training corpora of InternLM being Chinese supported. Comparing InternLM-7B, Qwen-7B, and LLaMA-7B with the same model parameters, the results of the first two models are significantly better than those of LLaMA, showing that the model using Chinese as the main corpus has significantly better capabilities for Chinese than the model using English as the main corpus. The ranking of all models is shown in Figure~\ref{fig:evaluation-all}.

In order to align with the level of human fundamental knowledge, we define 60 as the baseline of K-12 subjects section (the baseline for passing exams) and 68 as the baseline of common sense (the average scores test by human testers in common sense section). 

\textbf{Effect of model size. }
We observed models of different sizes, InternLM-123B, InternLM-70B, and InternLM-7B (they use corpora of similar size), LLaMA-7B, and LLaMA-13B pretrained on data of 1T tokens. From Figure~\ref{fig:evaluation-all} we can observe that the larger the model parameters, obviously the better accuracy.

\textbf{Question types and model capabilities. }
In the evaluation of common sense section, the scores of each model in the reasoning questions are significantly lower than the memory questions (taking InternLM-123B as an example, 62.28 vs. 86.06 <weighted average>). A paired sample t-test was performed on each model on reasoning questions and memory questions. The average score of reasoning questions was 28.69 points lower than that of memory questions (\emph{t} = -10.11, \emph{p} < 0.001), indicating that the knowledge reasoning capabilities of these models were significantly weaker than the memory-based knowledge capabilities.

\textbf{Circular. }
After using CircularEval, each model has a different degree of point drop, among which ChatGLM has the most serious point drop. This shows that although the model correctly selected the questions, it does not mean that it has indeed learned the relevant knowledge, and there is the possibility of coincidence. For models with lower accuracy, there are more drop points after CircularEval, and the two pairs show a negative correlation (\emph{r} = -0.47, \emph{p} < 0.001). This shows that in these data, the results of random guessing account for a higher proportion, so the accuracy is significantly reduced after using CircularEval. 

In common sense evaluation data, the drop point of the reasoning questions after using CircularEval is significantly higher than that of the memory questions (using Wilcoxon signed-rank test, \emph{z} = -2.201, \emph{p} = 0.018), which shows that the model randomly guesses the reasoning questions. The proportion of results is relatively high, again proving that the reasoning ability of each model is weak. 

\textbf{Hard examples. }
Following the macro analysis, we proceeded with a micro-level examination of the model results. We identified failure cases common to most models and discovered challenging examples that most ordinary people could answer correctly, but models often struggled to handle accurately. Examples are depicted in Appendix~\ref{sec:appendixB}.

\subsection{Conclusion}

In this paper, we present FoundaBench, a comprehensive benchmark for evaluating the foundational knowledge capabilities of large language models across common sense and K-12 educational subjects. This benchmark, enriched by a taxonomy spanning five main categories, 18 subcategories, 35 subjects, and 150 knowledge points, is upheld by rigorous multidimensional quality standards and manual quality control to ensure alignment with human foundational knowledge. Through the application of various evaluation methods, including zero-shot, few-shot, and circular comparisons on 14 diverse models, we have established a baseline to assess the foundational knowledge capabilities of LLMs.

\section{Limitations}
This study is based on Chinese scenarios to set a strategy for curating the dataset and designing the evaluation of the fundamental knowledge. However, the same method could be used to curate the evaluation dataset for other scenarios. 

\bibliography{FoundaBench}

\appendix
\section{Breakdown of model performance}
\label{sec:appendixA}
Table~\ref{tab:evaluation1-circular} and Table~\ref{tab:evaluation2-circular} show the detailed breakdown of models' accuracy per subject of common sense and K-12 section in zero-shot settings.

\begin{table*}[ht!]
    \centering
    \resizebox{0.99\textwidth}{!}{%
    \begin{tabular}
    { 
    cccccccc}
    \hline
    \textbf{Subjects} & \textbf{InternLM-123B} & \textbf{InternLM-70B} & \textbf{GPT4} & \textbf{GPT3.5-turbo} & \textbf{Qwen-14B} & \textbf{Baichuan2-13B} & \textbf{ChatGLM3-6B}\\
    \hline
    All & 65.89 & 67.11 & 55.74 & 39.7 & 52.79 & 47.72 & 26.8 \\
    \hline
    Reasoning & 43.94 & 49.48 & 26.99 & 20.42 & 25.26 & 22.49 & 7.96 \\
    Sci.\&Tech. & 85.42 & 81.25 & 80.21 & 65.63 & 64.58 & 66.67 & 38.54 \\
    Society & 75.78 & 75 & 65.63 & 42.97 & 62.5 & 56.25 & 28.12 \\
    Life & 61.26 & 65.77 & 58.56 & 43.24 & 47.75 & 46.85 & 28.83 \\
    Culture & 72.35 & 71.97 & 64.39 & 37.88 & 67.42 & 54.55 & 33.33 \\
    Art & 88.89 & 79.63 & 88.89 & 59.26 & 81.48 & 68.52 & 51.85 \\
    Ent. & 86.05 & 88.37 & 62.79 & 79.07 & 69.77 & 83.72 & 34.88 \\
    \hline
    \end{tabular}
    }
    \caption{Evaluation results by category in circular settings}
    \label{tab:evaluation1-circular}
\end{table*}

\begin{table*}[ht!]
\centering
\resizebox{0.99\textwidth}{!}{%
\begin{tabular}{
ccccccccc}
\hline
\textbf{Subjects} & \textbf{Education Phase} &\textbf{InternLM-123B} & \textbf{InternLM-70B} & \textbf{GPT4} & \textbf{GPT3.5-turbo} & \textbf{Qwen-14B} & \textbf{Baichuan2-13B} & \textbf{ChatGLM3-6B}\\
\hline
All &   & 50.58 & 45.35 & 47.63 & 18.64 & 40.31 & 23.35 & 11.88 \\
\hline
Chinese & Primary & 46.88 & 43.75 & 30 & 9.38 & 51.04 & 22.92 & 11.46 \\
& Middle & 52.08 & 38.54 & 16 & 2.08 & 63.54 & 18.75 & 5.21 \\
& High & 48.48 & 29.29 & 11 & 5.05 & 51.52 & 7.07 & 6.06 \\
\hline
Math & Primary & 20 & 17.89 & 43 & 11.58 & 37.89 & 10.53 & 7.37 \\
& Middle & 24 & 21 & 46 & 11 & 31 & 7 & 3 \\
& High & 17 & 11 & 36 & 2 & 11 & 3 & 1 \\
\hline
English & Primary & 65 & 55 & 63 & 35 & 69 & 26 & 6 \\
& Middle & 55 & 43 & 51 & 28 & 56 & 17 & 8 \\
& High & 69 & 51 & 62 & 36 & 59 & 26 & 18 \\
\hline
Physics & Middle & 55 & 42 & 56 & 14 & 28 & 18 & 14 \\
& High & 38 & 22 & 47 & 10 & 12 & 1 & 6 \\
\hline
Chemistry & Middle & 65.66 & 57.58 & 54 & 16.16 & 25.25 & 28.28 & 25.25 \\
& High & 20 & 13 & 17 & 5 & 8 & 2 & 3 \\
\hline
Biology & Middle & 63 & 64 & 66 & 34 & 36 & 34 & 20 \\
& High & 40 & 39 & 40 & 16 & 24 & 24 & 8 \\
\hline
Geography & Middle & 55 & 60 & 62 & 29 & 48 & 40 & 23 \\
& High & 52 & 48 & 53 & 21 & 39 & 24 & 12 \\
\hline
History & Middle & 75 & 78 & 58 & 29 & 58 & 43 & 24 \\
& High & 62 & 62 & 59 & 19 & 53 & 33 & 15 \\
\hline
Politics & Middle & 73.03 & 74.16 & 67 & 15.73 & 51.69 & 47.19 & 17.98 \\
& High & 56 & 51 & 36 & 13 & 62 & 37 & 12 \\
\hline
Info Tech. & High & 60.42 & 65.63 & 61 & 26.26 & 29.17 & 33.33 & 15.63 \\
\hline
General Tech. & High & 47.47 & 49.49 & 56 & 29 & 33.33 & 23.23 & 8.08 \\
\hline
Science & Middle & 54 & 52 & 53 & 30.21 & 30 & 34 & 15 \\
\hline
\end{tabular}
}
\caption{Evaluation results by education phase in circular settings}
\label{tab:evaluation2-circular}
\end{table*}

\section{Hard examples for LLMs}
\label{sec:appendixB}

The figures below show some hard examples that we found that most ordinary people could answer correctly, but models often struggled to handle accurately.

\textbf{Example 1: Correspondence reasoning questions }

\begin{figure*}
    \centering
    \includegraphics[width=0.8\linewidth]{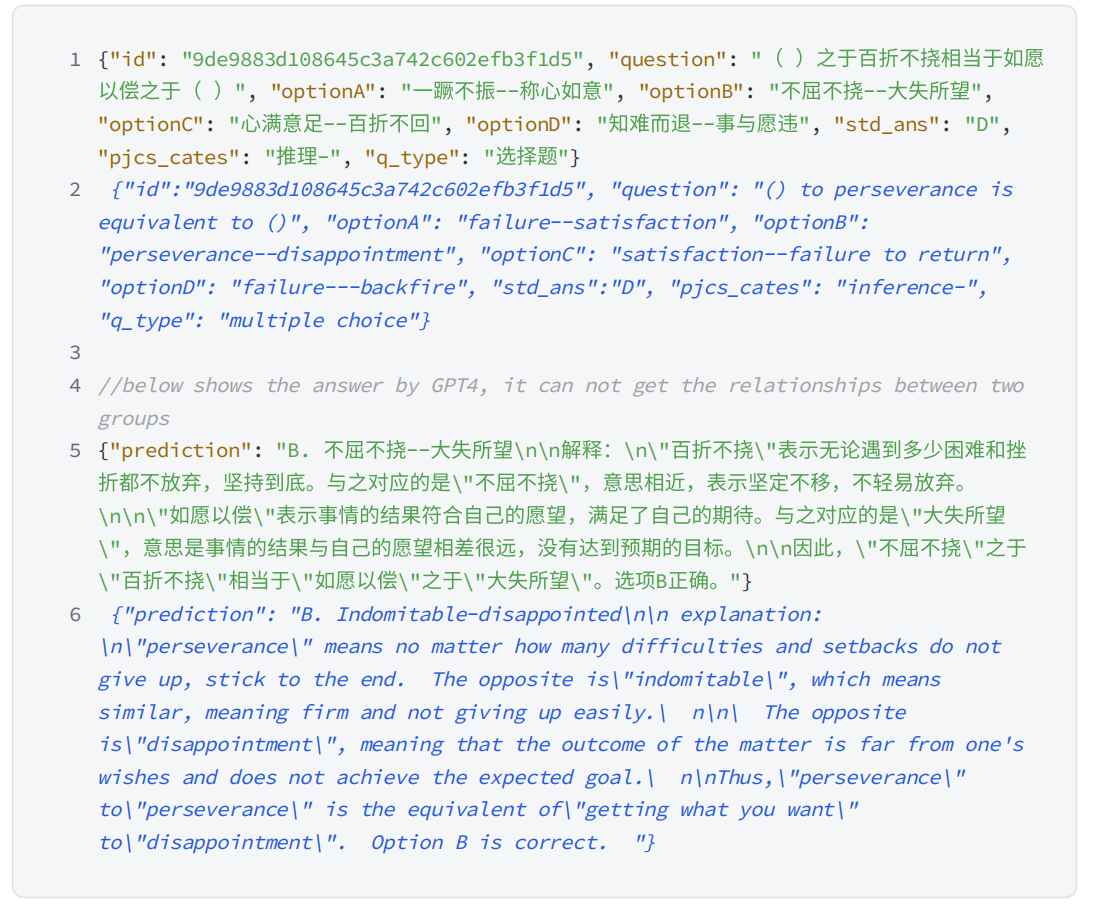}
    \caption{Hard example for correspondence reasoning question}
    \label{fig:example1}
\end{figure*}

For the given question, the human accuracy rate was 63. 33\%, while the LLMs had a success rate of 0/11, and none of them provided the correct answer. Most models chose 'B' as the answer. One of the examples is shown in Figure~\ref{fig:example1}.

\textbf{Example 2: Time calculation reasoning questions }

For the given question, the human accuracy rate was 80\%, while the LLMs had a success rate of 1/11, and only GPT4 gets the correct answer. One of the examples is shown in Figure~\ref{fig:example1}.

\begin{figure*}
    \centering
    \includegraphics[width=0.8\linewidth]{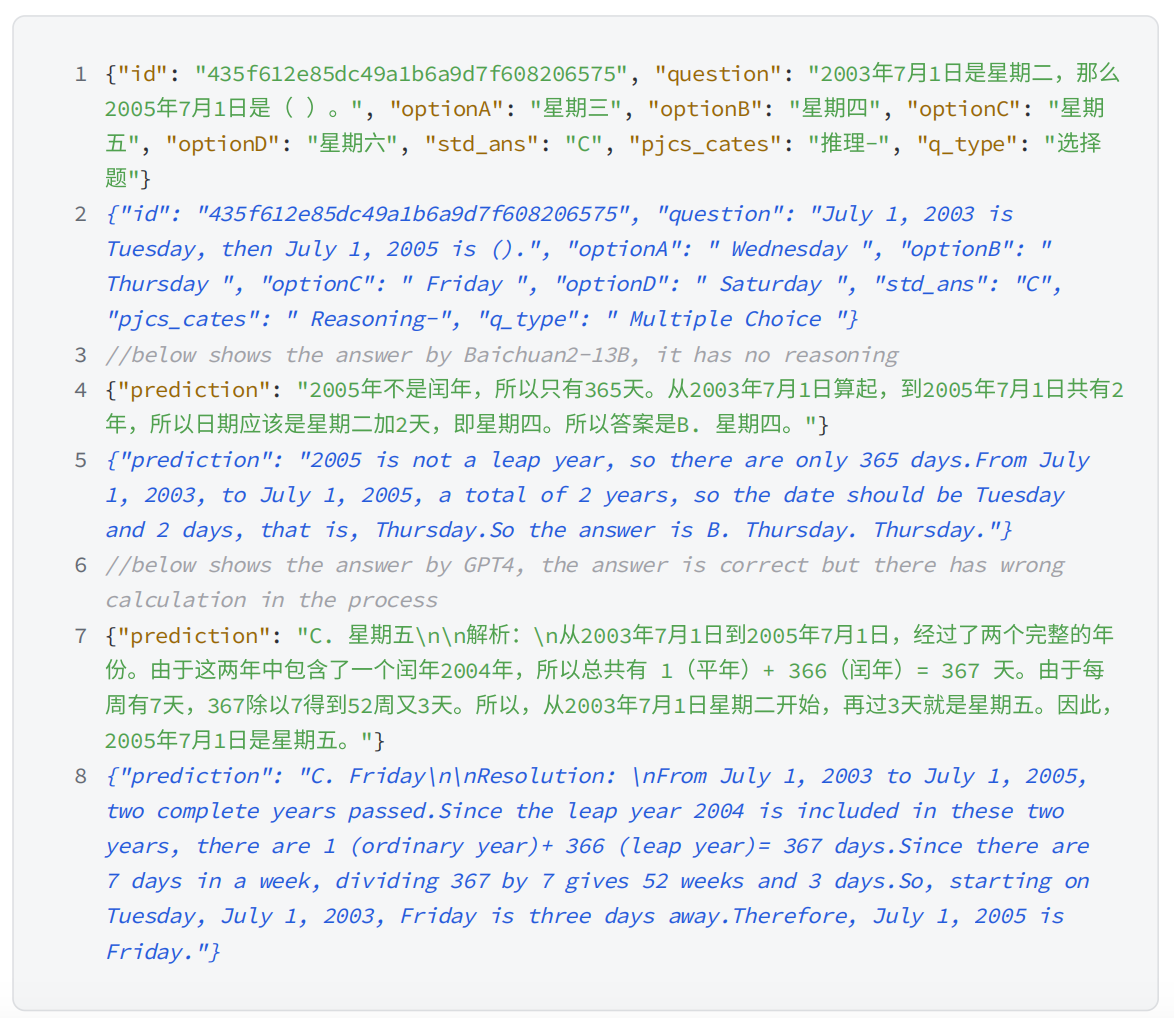}
    \caption{Hard example for time calculation reasoning question }
    \label{fig:example3}
\end{figure*}

\textbf{Example 3: numerical reasoning questions }

\begin{figure*}
    \centering
    \includegraphics[width=0.8\linewidth]{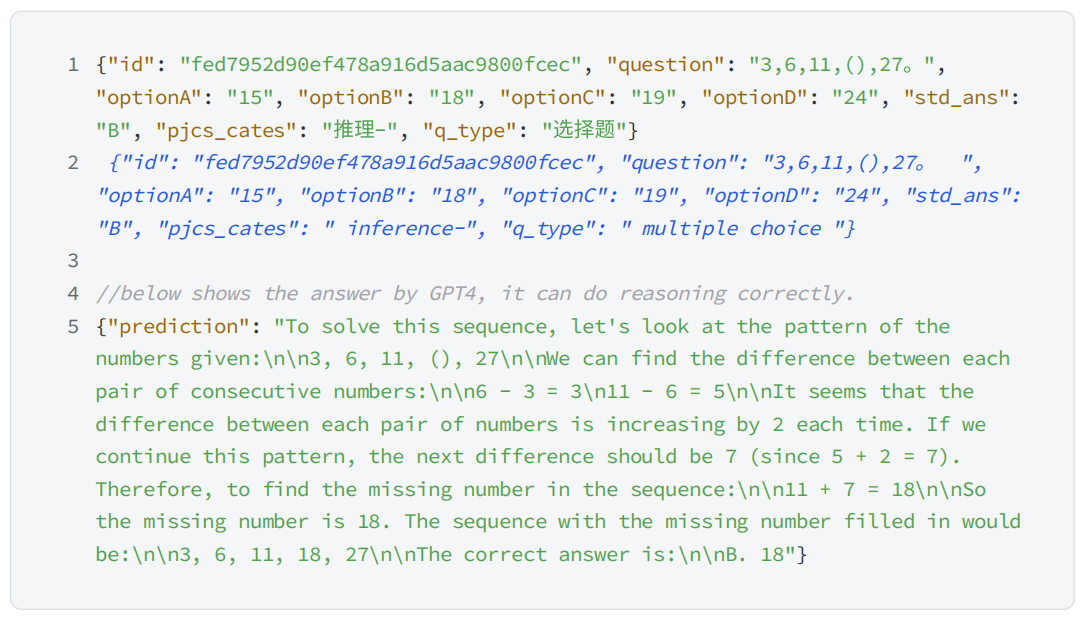}
    \caption{Hard example for numerical reasoning question }
    \label{fig:example2}
\end{figure*}

Example in Figure~\ref{fig:example2}, the human accuracy rate was 86.66\%, while LLMs had a success rate of 1/11, and only GPT4 gets the correct answer. 

\textbf{Example 4: questions for Chinese culture }

\begin{figure*}
    \centering
    \includegraphics[width=0.8\linewidth]{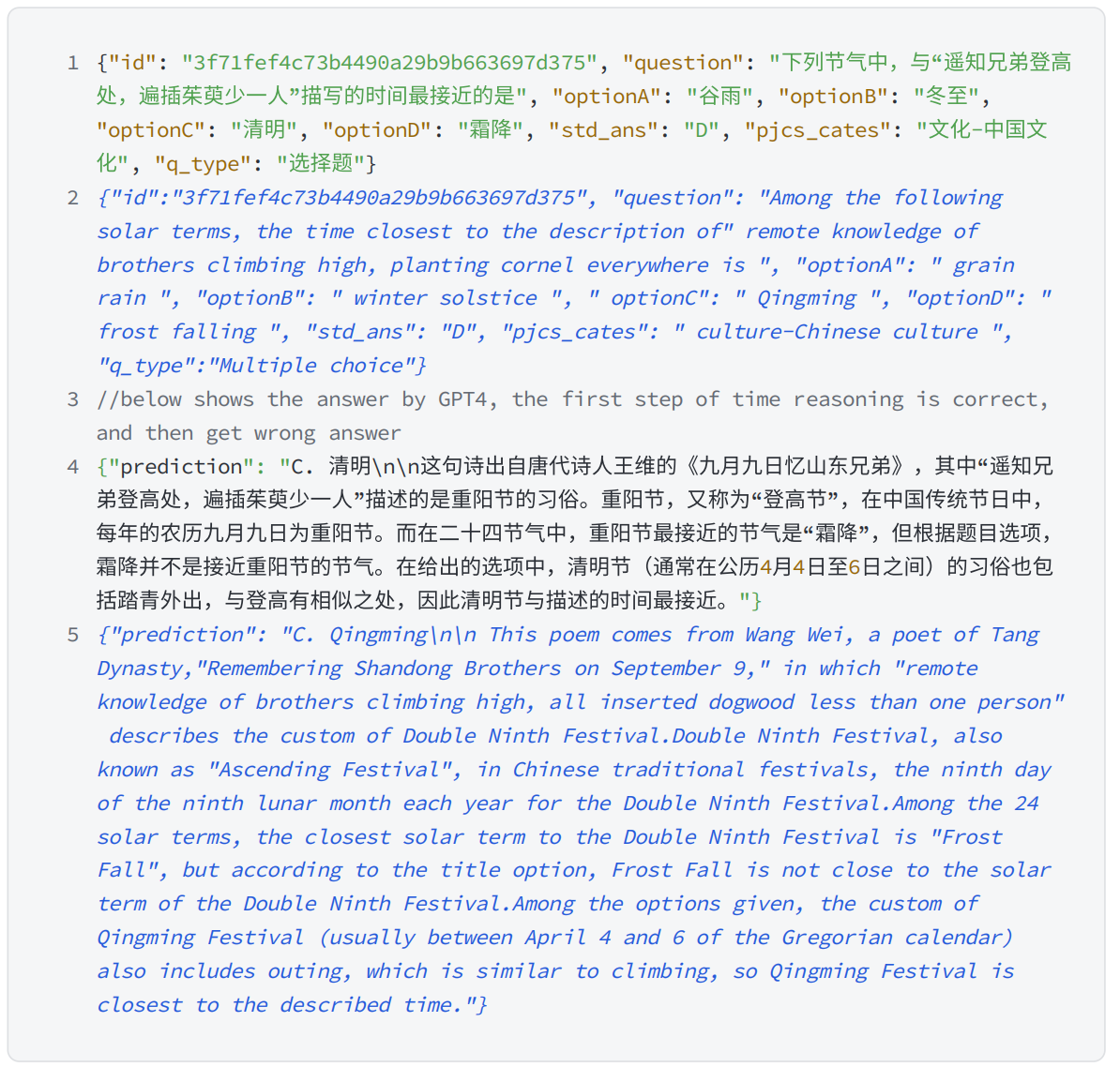}
    \caption{Hard example for Chinese culture }
    \label{fig:example4}
\end{figure*}

Example in Figure~\ref{fig:example4}, the human accuracy rate was 53.33\%, while LLMs had a success rate of 0/11, and none of them provided the correct answer. Most models chose 'C' as the answer. 

\end{document}